\documentclass[letterpaper, 10 pt, conference]{ieeeconf}  % Comment this line out if you need a4paper
\IEEEoverridecommandlockouts
\usepackage{cite}
\usepackage{balance}
\usepackage{amsmath,amssymb,amsfonts}
\usepackage{subfigure}
\usepackage{graphicx}
\usepackage{comment}
\usepackage{textcomp}
\usepackage{multirow} 
\usepackage{xcolor}
\usepackage{float}
\usepackage{tabularray}
\usepackage{hyperref}
\usepackage{array}
\usepackage{booktabs}
\usepackage{verbatim} 
\usepackage{makecell} % For line breaks in table cells
\usepackage{booktabs} % For better table lines
\usepackage{algorithm}
\usepackage{algpseudocode}
\usepackage{bm} % For bold math symbols
\newcommand{\ind}{\mathbb{1}}
\makeatletter
\usepackage[%
    top=0.78in,       % 上边距
    bottom=0.78in,    % 下边距
    left=0.78in,      % 左边距
    right=0.78in      % 右边距
]{geometry}

\newcommand{\Rmnum}[1]{\expandafter\@slowromancap\romannumeral #1@}

\makeatother
\def\BibTeX{{\rm B\kern-.05em{\sc i\kern-.025em b}\kern-.08em
    T\kern-.1667em\lower.7ex\hbox{E}\kern-.125emX}}

\title{\LARGE \bf Vision-Based Reasoning with Topology-Encoded Graphs for Anatomical Path Disambiguation in Robot-Assisted Endovascular Navigation}

\author{Jiyuan Zhao$^{1}$, Zhengyu Shi$^{1}$, Wentong Tian$^{1}$, Tianliang Yao$^{2}$, Dong Liu$^{3}$, Tao Liu$^{3}$, Yizhe Wu$^{4}$, Peng Qi$^{1, 5, *}$
\thanks{This work has been accepted by IEEE ICRA 2026. Copyright may be transferred, after which this version may no longer be accessible.}
\thanks{This work is supported by the National Key Research and Development Program of China under Grant No. 2023YFB4705200, the National Natural Science Foundation of China under Grant No. 62273257, and the Open Project Fund of State Key Laboratory of Cardiovascular Diseases No.2024SKL-TJ002. \emph{(*Corresponding Author: Peng Qi, email: pqi@tongji.edu.cn)}.}% <-this % stops a space
\thanks{$^{1}$Department of Control Science and Engineering, College of Electronics and Information Engineering, and Shanghai Institute of Intelligent Science and Technology, Tongji University, Shanghai 200092, China;}%
\thanks{$^{2}$Department of Electronic Engineering, Faculty of Engineering, The Chinese University of Hong Kong, Hong Kong SAR 999077, China;}
\thanks{$^{3}$Shanghai Operation Robot Co., Ltd., Shanghai 201318, China;}
\thanks{$^{4}$Department of Cardiology, Zhongshan Hospital, Fudan University, Shanghai Institute of Cardiovascular Diseases, National Clinical Research Center for Interventional Medicine, Shanghai 200032, China;}
\thanks{$^{5}$State Key Laboratory of Cardiovascular Diseases and Medical Innovation Center, Shanghai East Hospital, School of Medicine, Tongji University 200092, Shanghai, China.}
}

\begin{document}

\maketitle 
\pagestyle{empty}  % no page number for the second and the later pages
\thispagestyle{empty} % no page number for the first page

\begin{abstract}
Robotic-assisted percutaneous coronary intervention (PCI) is constrained by the inherent limitations of 2D Digital Subtraction Angiography (DSA). Unlike physicians, who can directly manipulate guidewires and integrate tactile feedback with their prior anatomical knowledge, teleoperated robotic systems must rely solely on 2D projections. This mode of operation, simultaneously lacking spatial context and tactile sensation, may give rise to projection-induced ambiguities at vascular bifurcations. To address this challenge, we propose a two-stage framework (SCAR-UNet-GAT) for real-time robotic path planning. In the first stage, SCAR-UNet, a spatial-coordinate-attention-regularized U-Net, is employed for accurate coronary vessel segmentation. The integration of multi-level attention mechanisms enhances the delineation of thin, tortuous vessels and improves robustness against imaging noise. From the resulting binary masks, vessel centerlines and bifurcation points are extracted, and geometric descriptors (e.g., branch diameter, intersection angles) are fused with local DSA patches to construct node features. In the second stage, a Graph Attention Network (GAT) reasons over the vessel graph to identify anatomically consistent and clinically feasible trajectories, effectively distinguishing true bifurcations from projection-induced false crossings. On a clinical DSA dataset, SCAR-UNet achieved a Dice coefficient of 93.1\%. For path disambiguation, the proposed GAT-based method attained a success rate of 95.0\% and a target-arrival success rate of 90.0\%, substantially outperforming conventional shortest-path planning (60.0\% and 55.0\%) and heuristic-based planning (75.0\% and 70.0\%). Validation on a robotic platform further confirmed the practical feasibility and robustness of the proposed framework.

\end{abstract}

\section{Introduction}
\begin{figure}[t]
\centering
\includegraphics[width=0.9\linewidth]{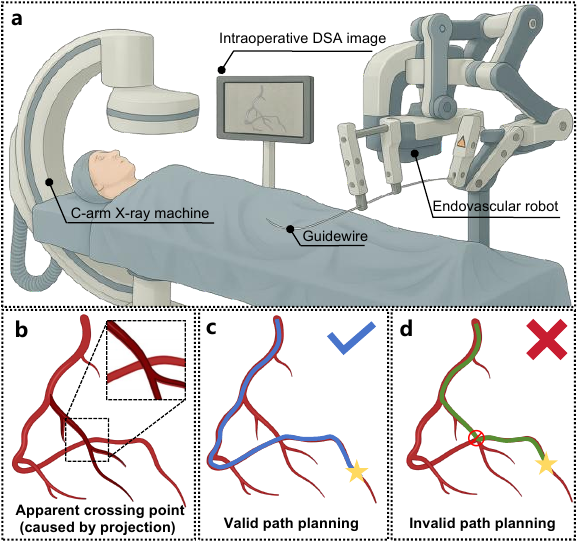}
\caption{The robotic intervention procedure relies on real-time 2D DSA images acquired by a C-arm X-ray system for path planning, providing precise navigation guidance for the robotic operation. (a) The intraoperative setup shows the endovascular robot operating under DSA guidance. (b) An apparent crossing point appears in the 2D projection when vessels at different depths visually overlap, which may be misinterpreted as a true bifurcation. (c) Correct path planning avoids such projection-induced crossings, while (d) invalid planning erroneously treats them as real intersections, potentially leading to navigation errors. The schematic was created using BioRender (\url{https://biorender.com}).}
\label{fig:Illustration}
\end{figure}

Percutaneous coronary intervention (PCI) depends on the physician’s ability to navigate tortuous vascular structures under two-dimensional (2D) Digital Subtraction Angiography (DSA) guidance \cite{yao2023enhancing}. Physicians can partially compensate for the projectional limitations of DSA by integrating tactile feedback and prior anatomical knowledge. Nevertheless, vessel overlap and perspective ambiguity remain significant problems, particularly at bifurcations and intersections \cite{yao2025realtip,yao2026advancing}. Robotic-assisted endovascular systems have been introduced to reduce dependence on individual operator expertise and to standardize procedures. However, in the absence of perceptual and cognitive feedback, these systems depend exclusively on 2D DSA, which magnifies anatomical ambiguities and complicates the development of reliable path planning algorithms \cite{adhami2003optimal,yang2017medical}.

Although these limitations present substantial challenges, robotic-assisted interventions offer clear potential to improve the accuracy, reproducibility, and minimally invasive nature of PCI \cite{yao2025sim2real, yao2025sim4endor}. Achieving this potential requires overcoming a decisive bottleneck, namely the design of navigation strategies capable of generating anatomically correct and clinically feasible guidewire trajectories in real time \cite{vakharia2019effect,yao2026advancing,delgado2021use,yao2025multi}. Addressing this challenge is essential both for ensuring patient safety and for enabling the widespread clinical adoption of robotic PCI.

Path planning for endovascular navigation has been investigated through both hemodynamic modeling and medical imaging approaches \cite{devineni2022diagnostic,li2026machine,zhao2022surgical,li2026hierarchical,yao2023enhancing,ravigopal2021automated}. Hemodynamic models provide insight into global circulation and functional assessment, but their reliance on specialized instrumentation and limited suitability for intraoperative use restricts real-time applicability. Imaging-based strategies have therefore gained wider adoption. Modalities such as intravascular ultrasound (IVUS) and computed tomography angiography (CTA) provide valuable anatomical information for planning, yet they are not typically available for continuous intraoperative guidance. In contrast, two-dimensional (2D) Digital Subtraction Angiography (DSA) serves as the primary intraoperative modality. PCI is conducted under real-time DSA guidance from vascular access to lesion treatment, and navigation decisions are made directly from these projections. Consequently, DSA constitutes the most consequential imaging source for practical path planning. Recent progress in deep learning has further strengthened DSA-based approaches, with segmentation networks such as U-Net variants achieving notable success in extracting vessels from noisy and highly complex angiographic images \cite{ronneberger2015u,xiao2018weighted,chen2021transunet}.

Despite these advances, a fundamental limitation remains. Most imaging-based navigation methods assume that the segmented topology obtained from 2D data accurately reflects the underlying three-dimensional (3D) anatomy. In reality, this assumption does not hold, as the projective nature of DSA inevitably introduces apparent overlaps and false crossings that are unrelated to true bifurcations (see Fig.~\ref{fig:Illustration}). These ambiguities distort the vessel graph representation and can misguide conventional path planning algorithms, resulting in navigation trajectories that fail to respect vascular physiology or clinical safety.

In response, researchers have explored methods that incorporate richer geometric constraints or multi-view information. Frisken et al. proposed a constrained 3D reconstruction approach that combines vessel centerlines, radii, and contrast flow from biplane DSA to recover cerebral vascular structures \cite{frisken2025spatiotemporally}. Song et al. developed an iterative Perspective-n-Point (PnP) registration framework for aligning preoperative 3D vascular data with 2D intraoperative images in real time \cite{song2024iterative}. These techniques highlight the promise of augmenting 2D navigation with 3D consistency. Nevertheless, their clinical practicality remains limited due to specific imaging requirements, the computational overhead of reconstruction, and challenges in achieving reliable real-time performance.

To address this challenge, this study presents a topology‑aware pipeline for vascular path planning under 2D DSA. First, vessel centerlines are extracted from segmentation masks and key nodes such as bifurcations and projection‑induced crossings are identified. Next, for each node and connecting segment, geometric features such as vessel diameter, branch angles, and segment lengths, together with local image features, are aggregated into feature vectors. These descriptors define a vessel graph that is processed by a Graph Attention Network, abbreviated as GAT, to estimate path traversability and anatomical plausibility. The resulting scores suppress spurious 2D crossings and prioritize clinically consistent routes. By explicitly modeling vascular topology, the pipeline refines segmentation outputs and improves graph‑based planning, advancing toward more reliable robotic PCI.

The primary contributions of this work are as follows:
\begin{itemize}
    \item \textbf{SCAR-UNet-GAT Framework:} A unified framework integrating spatial-coordinate attention-based segmentation with topology-aware graph neural reasoning for vascular path planning under 2D DSA guidance.
    \item \textbf{Vessel Graph Construction:} A graph-construction strategy that incorporates geometric and image-derived node and edge features, enabling robust differentiation between true bifurcations and projection-induced crossings.
    \item \textbf{Topology-Aware Graph Reasoning:} A GAT-based inference module leveraging vessel topology, geometry, and localized DSA patches to reduce projection-induced ambiguities and achieve higher crossing disambiguation accuracy and target arrival rates than shortest-path and heuristic planners.
\end{itemize}

\begin{figure*}[t]
\centering
\includegraphics[width=0.86\linewidth]{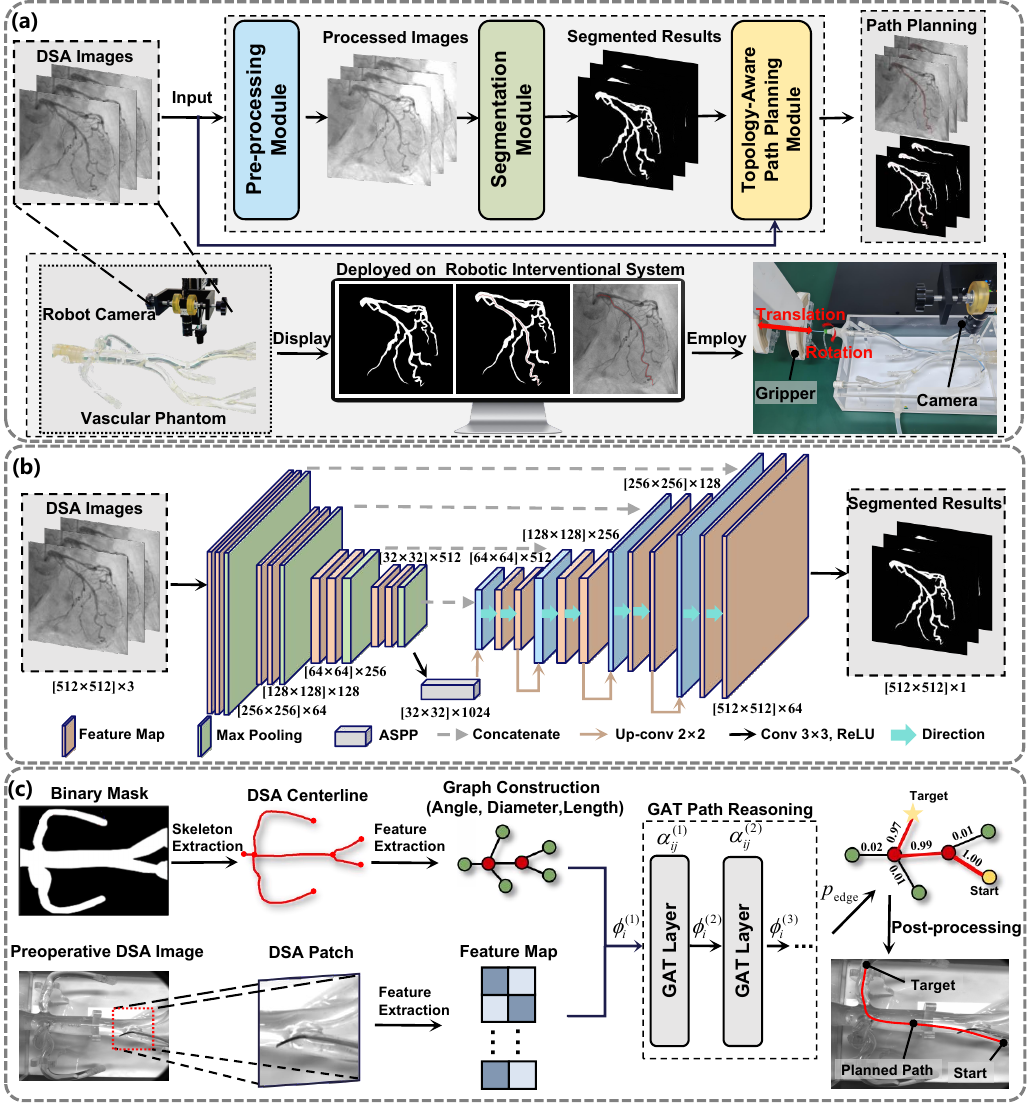}
\caption{Overview of the proposed SCAR-UNet-GAT framework for robotic vascular path planning. (a) The pipeline combines SCAR-UNet-based segmentation of DSA image sequences with topology-aware GAT-based path planning, and is deployed on a robotic interventional system to provide navigation guidance for intervention procedures. (b) The SCAR-UNet segmentation module uses an encoder-decoder architecture with spatial-coordinate attention, integrated with multiple attention mechanisms and ASPP, enabling accurate extraction of vascular masks from DSA images. (c) The topology-aware path planning module extracts vessel centerlines and crossing features from segmentation results, fuses image patch features at intersections, constructs a vessel graph, predicts path probabilities using GAT reasoning, and then undergoes post-processing to determine the planned route.}
\label{fig:OverallFramework}
\end{figure*}

\section{Methodology}
\subsection{System Overview}
The proposed SCAR-UNet-GAT, illustrated in Fig.~\ref{fig:OverallFramework}(a), delivers a unified perception-to-planning pipeline for topology-aware vascular route inference in robot-assisted endovascular navigation. The architecture comprises three stages: preprocessing, high-fidelity vessel segmentation, and graph-based reasoning for path selection.

The pipeline starts from 2D DSA acquisition, followed by preprocessing that enhances vascular contrast and boundary delineation. SCAR-UNet then produces binary vessel masks with reliable depiction of thin and tortuous branches. From these masks, centerlines are extracted and key nodes, including bifurcations and projection-induced crossings, are detected to construct a vessel graph by linking node pairs along unique centerline segments. Geometric descriptors such as local diameter, branch angles, and segment length, together with appearance features from localized DSA patches, are encoded as node and edge attributes. A Graph Attention Network, abbreviated GAT, estimates edge traversability and anatomical plausibility to disambiguate apparent crossings and selects an anatomically consistent route for guidewire navigation. The workflow is deployed on a robotic platform as a preoperative planning tool. Algorithm~\ref{alg:scarunetgnn} outlines the procedure.

\begin{algorithm}[!htbp]
\caption{SCAR-UNet-GAT Path Planning Pipeline}
\label{alg:scarunetgnn}
\begin{algorithmic}[1]
\Require DSA image $\bm{x} \in \mathbb{R}^{512 \times 512}$ \hfill \Comment{Input image} 
\Statex \hspace{2.2em} source node $i_{\text{src}}$, target node $i_{\text{tgt}}$ \hfill \Comment{Endpoints}

\Ensure Vessel path $\mathcal{P}$  \Comment{Output path}
\Statex
\State $\bm{z} \gets \text{SCAREncoder}(\bm{x})$  \Comment{SCAR blocks, encoder}
\State $\bm{c} \gets \text{ASPP}(\bm{z})$  \Comment{Atrous spatial pyramid pooling}
\State $\bm{d} \gets \text{SCARDecoder}(\bm{c}, \bm{z})$  \Comment{SCAR blocks, decoder}
\State $\hat{\bm{y}} \gets \sigma(\text{Conv}_{1\times 1}(\bm{d}))$  \Comment{Probability map}
\State $\bm{m} \gets \ind\{ \hat{\bm{y}} \geq \tau \}$  \Comment{Binary vessel mask}

\State $\mathcal{C} \gets \text{Skeletonize}(\bm{m})$  \Comment{Centerlines}
\State $\mathcal{N} \gets \text{KeyNodeDetect}(\mathcal{C})$  \Comment{Bifurcations and intersections}
\State $\phi_i \gets \text{FeatExtract}(\mathcal{N},\,\mathcal{C},\, \bm{m},\, \bm{x})$  \Comment{Node and segment features}

\State $\mathcal{G} \gets \text{GraphConstruct}(\mathcal{N},\, \{\phi_i\})$  \Comment{Vessel graph}
\State $p_{\text{edge}} \gets \text{GAT}(\mathcal{G})$  \Comment{Edge traversability scores}
\State $\mathcal{P} \gets \text{PathPostprocess}(\mathcal{G},\, i_{\text{src}},\, i_{\text{tgt}},\, p_{\text{edge}})$  \Comment{Shortest path, smoothing}
\State \Return $\mathcal{P}$
\Statex
\Function{SCARBlock}{$\bm{f}$}
    \State $\bm{f} \gets \text{CoordAttention}(\bm{f})$  \Comment{Coordinate attention}
    \State $\bm{f} \gets \text{SimAM}(\bm{f})$  \Comment{Parameter-free attention}
    \State $\bm{f} \gets \text{SE}(\bm{f})$  \Comment{Channel reweighting}
    \State \Return $\bm{f}$
\EndFunction
\end{algorithmic}
\end{algorithm}

\subsection{Problem Formulation}
Let $\bm{x}\in\mathbb{R}^{512\times512}$ be a 2D DSA image defined on pixel domain $\Omega$ with spacing $s=(s_x,s_y)$. The segmentation module outputs a vessel mask $m:\Omega\to\{0,1\}$ and the skeletonization yields a centerline set $\mathcal{C}$. A vascular graph $\mathcal{G}=(\mathcal{N},\mathcal{E})$ is constructed from $\mathcal{C}$, where $\mathcal{N}$ contains endpoints and bifurcations, and $\mathcal{E}$ connects node pairs linked by a unique centerline segment. Each node $i\in\mathcal{N}$ is associated with a feature vector $\phi_i\in\mathbb{R}^D$ that concatenates geometric descriptors and multi-scale appearance embeddings as defined in the feature extraction stage.

The planning objective is to compute an anatomically plausible centerline route between a specified source node $i_{\text{src}}\in\mathcal{N}$ and target node $i_{\text{tgt}}\in\mathcal{N}$. Let $\Pi_{i_{\text{src}}\to i_{\text{tgt}}}(\mathcal{G})$ denote the set of simple paths on $\mathcal{G}$ that connect $i_{\text{src}}$ and $i_{\text{tgt}}$ without cycles. A Graph Attention Network maps node features and graph structure to edge likelihoods $p_{ij}\in(0,1)$ for $(i,j)\in\mathcal{E}$, which quantify edge traversability and anatomical plausibility under the given projection.

Path selection is posed as a discrete optimization over the graph:

\begin{equation}
    \mathcal{P}^{\ast} \;=\; \arg\max_{\mathcal{P}\in\Pi_{i_{\text{src}}\to i_{\text{tgt}}}(\mathcal{G})}
\; \sum_{(i,j)\in\mathcal{P}} \log p_{ij},
\end{equation}

which favors chains of high-likelihood edges and enforces topological consistency by construction. The problem is equivalently solved as a shortest-path search on edge costs $w_{ij}=-\log p_{ij}$. Optional geometric priors, such as curvature or diameter regularization defined from $\theta_{ij}$ and $d_{ij}$, can be incorporated as additive costs on edges without changing the path-search formulation.

The output is a centerline path $\mathcal{P}^{\ast}$ represented as an ordered edge sequence on $\mathcal{G}$. A smoothing step produces a continuous trajectory aligned with the underlying skeleton to support guidewire navigation.

\subsection{High-Fidelity Vessel Segmentation for Geometric Feature Extraction}
Accurate vessel segmentation in 2D DSA is challenged by low contrast, heavy imaging noise, and severe class imbalance between vessel and background \cite{yao2025real}. Overlapping bone structures and faint peripheral signals further obscure boundaries, which increases the risk of broken centerlines and missed small branches. These factors motivate an architecture with strong representational capacity and precise spatial localization.

To address this need, SCAR-UNet is adopted for vessel extraction, as shown in Fig.~\ref{fig:OverallFramework}(b). The input image $\bm{x} \in \mathbb{R}^{512\times512}$ is processed by an encoder built from SCARBlocks, denoted $\mathcal{E}_{\text{SCAR}}$, yielding a multiscale representation $\bm{z} = \mathcal{E}_{\text{SCAR}}(\bm{x})$ with shape $\mathbb{R}^{H' \times W' \times D}$. Each SCARBlock applies CoordAttention, SimAM, and SE to enhance spatial sensitivity, emphasize informative regions, and reweight channels. This combination preserves fine vessel geometry, highlights relevant pixels, and suppresses noise, which improves delineation of thin and tortuous segments.

The bottleneck features are refined by an atrous spatial pyramid pooling module $\mathcal{A}_{\text{ASPP}}$, producing a context-enriched tensor $\bm{c} = \mathcal{A}_{\text{ASPP}}(\bm{z}) \in \mathbb{R}^{H' \times W' \times 1024}$. A decoder $\mathcal{D}_{\text{SCAR}}$ with skip connections hierarchically upsamples $\bm{c}$ and fuses encoder features to reconstruct high-resolution maps $\bm{d} = \mathcal{D}_{\text{SCAR}}(\bm{c}, \bm{z}) \in \mathbb{R}^{512 \times 512 \times C_d}$. A $1\times 1$ projection with sigmoid activation yields the vessel probability map $\hat{\bm{y}} = \sigma(\mathcal{C}_{1\times1}(\bm{d})) \in [0,1]^{512\times512}$. To facilitate optimization and provide richer supervision, an auxiliary prediction $\hat{\bm{y}}_{\text{aux}}$ is generated from intermediate decoder features.

Vascular segmentation exhibits extreme foreground sparsity. To mitigate bias toward background prediction while capturing thin structures, a hybrid loss combines soft Dice loss with Focal Tversky loss for both the main and auxiliary outputs:
\begin{equation}
\begin{aligned}
    \mathcal{L}_{\text{total}} =
    \omega_1\,\mathcal{L}_{\text{Dice}}(\bm{y}, \hat{\bm{y}})
  + \omega_2\,\mathcal{L}_{\text{FT}}(\bm{y}, \hat{\bm{y}})
  + \omega_3\,\mathcal{L}_{\text{Dice}}(\bm{y}, \hat{\bm{y}}_{\text{aux}})\\
  + \omega_4\,\mathcal{L}_{\text{FT}}(\bm{y}, \hat{\bm{y}}_{\text{aux}}),
\end{aligned}
\end{equation}
where $\bm{y} \in \{0,1\}^{512\times512}$ is the ground-truth mask and $\omega_1,\omega_2,\omega_3,\omega_4$ are nonnegative weights. The soft Dice loss is
\begin{equation}
\mathcal{L}_{\text{Dice}}(\bm{y}, \hat{\bm{y}})
= 1 - \frac{2 \sum_{i} \hat{y}_i y_i + \epsilon}
           {\sum_{i} \hat{y}_i^2 + \sum_{i} y_i^2 + \epsilon},
\label{eq:dice_loss}
\end{equation}
and the Focal Tversky loss is defined using
$\mathrm{TP}=\sum_i \hat{y}_i y_i$, 
$\mathrm{FP}=\sum_i \hat{y}_i (1-y_i)$, and 
$\mathrm{FN}=\sum_i (1-\hat{y}_i) y_i$ as
\begin{equation}
\mathcal{L}_{\text{FT}}(\bm{y}, \hat{\bm{y}})
= \left( 1 - \frac{\mathrm{TP} + \epsilon}{\mathrm{TP} + \alpha\,\mathrm{FP} + \beta\,\mathrm{FN} + \epsilon} \right)^{\gamma},
\label{eq:focal_tversky}
\end{equation}
with $\alpha$, $\beta$, and $\gamma$ controlling the penalties for false positives and false negatives, and $\epsilon$ ensuring numerical stability. Joint optimization of these complementary terms improves overlap on thin vessels, emphasizes hard pixels, and stabilizes training under severe class imbalance. The resulting probability maps support reliable centerline extraction and geometric feature computation for the topology-aware planning module.

\subsection{Centerline Extraction and Feature Encoding for Vascular Graph Construction}
Given a binary vessel mask $m:\Omega\to\{0,1\}$, a morphological closing operator with structuring element $K$ is first applied for denoising and gap filling, yielding $\tilde{m}=\mathcal{M}(m;K)$. Skeletonization $\mathcal{S}(\cdot)$ produces the centerline set
$\mathcal{C}=\{\,q\in\Omega \mid \mathcal{S}(\tilde{m})(q)=1\,\}$.
All skeleton pixels are retained at this stage. To suppress topological artifacts, short spurs are pruned: a path $P$ is removed if its arc length $l(P)$ is below $L_{\min}$ and every internal node has degree $2$. Key nodes $\mathcal{N}$ are defined as endpoints with $\deg(v)=1$ and bifurcations with $\deg(v)\geq 3$. Let $E$ denote the set of edges. Two nodes $i,j\in\mathcal{N}$ are connected by an undirected edge $e_{ij}\in E$ if a unique skeleton path exists between them without intermediate key nodes.

For each edge $e_{ij}$, order the pixels as $\{p_k\}_{k=1}^{|e_{ij}|}$ and define tangent steps $u_k=p_{k+1}-p_k$. The mean turning angle is
\begin{equation}
\theta_{ij}=\frac{1}{|e_{ij}|-2}\sum_{k=1}^{|e_{ij}|-2}
\arccos\!\left(\frac{\langle u_k,u_{k+1}\rangle}{\|u_k\|_2\,\|u_{k+1}\|_2}\right),
\label{eq:turning_angle}
\end{equation}
where $u_k$ and $u_{k+1}$ are successive tangent vectors along the skeleton.

Let the pixel spacing be $s=(s_x,s_y)$ and define the metric matrix $M=\mathrm{diag}(s_x,s_y)$. The physical length of an edge is
\begin{equation}
l_{ij}=\sum_{k=1}^{|e_{ij}|-1}\|M\,u_k\|_2,
\label{eq:arc_length}
\end{equation}
which gives the discrete arc length in physical units. Vessel diameter is computed from the distance transform of $\tilde{m}$. Let $D_M(p)$ denote the radius at pixel $p$ measured under the metric $M$ (anisotropic distance transform). The edge-wise diameter is $d_{ij}=\frac{2}{|e_{ij}|}\sum_{p\in e_{ij}} D_M(p),$ which reduces to $d_{ij}=\tfrac{2}{|e_{ij}|}\sum_{p\in e_{ij}} D(p)\,s$ in the isotropic case with spacing $s$.

Node-level geometric descriptors aggregate statistics over incident edges. For node $i$, $\theta_i=\frac{1}{\deg(i)}\sum_{j:(i,j)\in E}\theta_{ij},\quad
d_i=\max_{j:(i,j)\in E} d_{ij},\quad
l_i=\frac{1}{\deg(i)}\sum_{j:(i,j)\in E} l_{ij}.$

Appearance cues are captured by multi-scale patches $\{P_i^{(s)}\in\mathbb{R}^{s\times s}\mid s\in\{32,64,96\}\}$ cropped from the original DSA image $x$ at node $i$. Patches are normalized to zero mean and unit variance and processed by a shared ResNet-18 backbone pretrained on ImageNet. Global pooling yields embeddings $f_i^{(32)}$, $f_i^{(64)}$, and $f_i^{(96)}$. The final node descriptor is $\phi_i=\operatorname{concat}\!\big(\theta_i,\, d_i,\, l_i,\, f_i^{(32)},\, f_i^{(64)},\, f_i^{(96)}\big)\in\mathbb{R}^D,$ which provides a robust representation for downstream graph reasoning and vascular path planning.
\subsection{Topology-Aware Path Planning with Graph Attention Networks}
For path inference, a Graph Attention Network (GAT) is adopted in place of Graph Convolutional Networks and graph transformer models. Unlike GCNs that weight neighbors uniformly, GAT learns adaptive attention coefficients that emphasize clinically relevant bifurcations and suppress projection-induced crossings. Compared with graph transformers, GAT is more parameter efficient and supports real-time navigation on the target platform. This choice enables explicit topology-aware reasoning by aggregating geometric and appearance cues from adjacent nodes and improves discrimination between true bifurcations and 2D crossings.

Given the vessel graph $\mathcal{G}=(\mathcal{N},\mathcal{E})$ with node features $\phi_i$ for $i\in\mathcal{N}$, attention coefficients $\alpha_{ij}$ for neighbors $j\in\mathcal{N}(i)$ are computed as
\begin{equation}
\alpha_{ij}
= \frac{\exp\!\left(\mathrm{LeakyReLU}\!\left(\mathbf{a}^\top\!\big[\mathbf{W}\phi_i \, \Vert \, \mathbf{W}\phi_j\big]\right)\right)}
{\sum_{k\in\mathcal{N}(i)} \exp\!\left(\mathrm{LeakyReLU}\!\left(\mathbf{a}^\top\!\big[\mathbf{W}\phi_i \, \Vert \, \mathbf{W}\phi_k\big]\right)\right)},
\label{eq:attention_coeff}
\end{equation}
where $\mathbf{W}$ is a learnable weight matrix, $\mathbf{a}$ is the attention vector, and $\Vert$ denotes concatenation. Node embeddings are then updated by attention-weighted aggregation $\tilde{\phi}_i = \mathrm{ReLU}\!\left(\sum_{j\in\mathcal{N}(i)} \alpha_{ij}\,\mathbf{W}\phi_j\right).$

Multi-head attention can be employed and combined by averaging or concatenation to improve stability.

Edge likelihoods are obtained with a bilinear scorer
\begin{equation}
p_{ij} = \mathrm{sigmoid}\!\left(\tilde{\phi}_i^\top \mathbf{W}_e \tilde{\phi}_j + b_e\right),
\label{eq:edge_probability}
\end{equation}
where $\mathbf{W}_e$ is a learnable matrix and $b_e$ is a bias term. The optimal route along the vessel centerlines is extracted by a shortest-path search on negative log-likelihoods $-\log p_{ij}$, which favors chains of high-probability edges. A smoothing step then removes jagged segments and enforces a continuous, topologically consistent trajectory for guidewire navigation.

\begin{figure*}[t]
\centering
\includegraphics[width=0.86\linewidth]{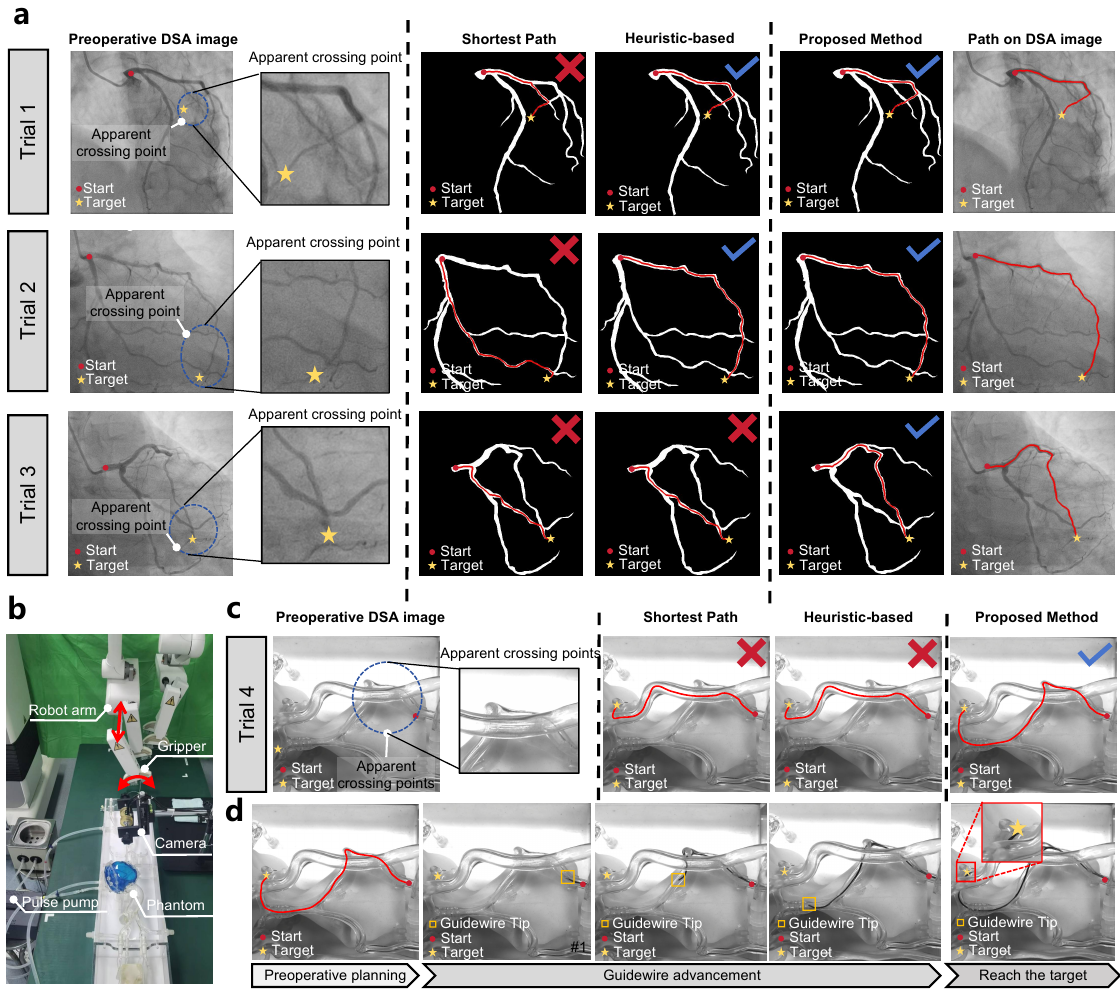}
\caption{Validation of Robotic Vascular Path Planning on Clinical Data and Robotic Systems. (a) Results on clinical DSA images, illustrating the details of apparent crossing points and the visual comparison among three methods: shortest-path, heuristic-based, and the proposed approach. (b) The robotic intervention platform, consisting of a robotic arm, gripper, monocular camera, pulse pump, and vascular phantom. The monocular camera provides DSA-like vascular images as inputs for the path planning system. (c) Results on vascular phantom images, further demonstrating the disambiguation of apparent crossing points and the comparative visualization of the three methods. (d) Validation on robotic systems, where YOLOv5 was employed for real-time guidewire tip detection and PID-based control was used to regulate advancement and rotation of the gripper, enabling the guidewire to follow the planned trajectory and successfully reach the target point.
}
\label{fig:Result}
\end{figure*}
\section{Experiments and Results}
\subsection{Experimental Setup}
The DSA dataset comprised anonymized coronary angiograms annotated by experienced cardiologists: 768 training images from 192 patients, 200 validation images from 73 patients, and 109 test images from 23 patients (single-channel $512\times512$), with no patient overlap; curation was performed in 3D Slicer. Experiments were conducted on clinical DSA images and on a vascular phantom platform from Shanghai Operation Robot Co., Ltd., which includes a robotic arm, gripper, pulsatile pump, and a monocular camera producing DSA-like inputs, as shown in Fig.~\ref{fig:Result}(b).

All models were implemented in PyTorch 2.7.0 with OpenCV 4.11.0.86 and run on an Intel Core i7-14700HX CPU with 32~GB RAM and an NVIDIA RTX~4060 GPU. The segmentation network was trained for 100 epochs with a batch size of 16 using warm-up and cosine annealing, an initial learning rate of $1\times10^{-4}$, and weight decay of $1\times10^{-4}$. The hybrid loss used $\omega_1=\omega_3=1.0$ and $\omega_2=\omega_4=0.5$; the Focal Tversky parameters were $\alpha=0.7$, $\beta=0.3$, and $\gamma=0.75$. In the GAT module, the LeakyReLU negative slope was set to $0.2$. Data augmentation applied random geometric and photometric transforms, including flips, $90^\circ$ rotations, brightness and contrast adjustment, Gaussian blur and noise, gamma correction, zoom, affine shift, elastic deformation, and crop-resize; each image–mask pair was augmented up to sixfold. For graph-based planning, bifurcations and endpoint connectivity were manually annotated by experienced clinicians to provide supervision.

\begin{table}[!htbp]
\centering
\caption{Comparative Analysis of Vessel Segmentation Performance}
\label{tab:seg}
\begin{tabular}{lcccc}
\toprule
\textbf{Method} & \textbf{Dice (\%)} & \textbf{Precision (\%)} & \textbf{Recall (\%)} & \textbf{FPS} \\
\midrule
U-Net            & 89.4 & 87.7 & 90.9 & 68 \\
Attention U-Net  & 90.5 & 89.0 & 92.0 & 61 \\
DeepLabv3+       & 90.8 & 89.1 & 92.0 & 62 \\
TransUNet        & 91.6 & 90.8 & 92.5 & 49 \\
Swin-UNet        & 92.0 & 91.2 & 93.1 & 45 \\
\textbf{SCAR-UNet} & \textbf{93.1} & \textbf{92.0} & \textbf{94.5} & 58 \\
\bottomrule
\end{tabular}
\end{table}

\begin{table}[!htbp]
\centering
\caption{Ablation study of SCAR-UNet on the coronary DSA dataset. Dice, Precision, and Recall are reported in \%; FPS measures inference speed.}
\label{tab:ablation}
\begin{tabular}{lcccc}
\toprule
\textbf{Model Variant}                & \textbf{Dice} & \textbf{Precision} & \textbf{Recall} & \textbf{FPS} \\
\midrule
w/o CoordAttention                    & 91.7 & 90.4 & 92.8 & 60 \\
w/o SimAM                             & 92.0 & 90.7 & 93.2 & 60 \\
w/o SE                                & 92.2 & 91.0 & 93.3 & 59 \\
w/o ASPP                              & 92.5 & 91.4 & 93.6 & 59 \\
w/o Hybrid Loss (Dice only)           & 92.4 & 91.2 & 93.4 & 58 \\
\textbf{Full SCAR-UNet}               & \textbf{93.1} & \textbf{92.0} & \textbf{94.5} & 58 \\
\bottomrule
\end{tabular}
\end{table}

\begin{table*}[!htbp]
\centering
\small
\caption{Comparison of Apparent Crossing Disambiguation and Target Arrival on DSA Images}
\label{tab:comparison}
\begin{tabular}{ccc}
\toprule
\textbf{Method} & \makecell[c]{\textbf{Disambiguation Success Rate} \\ \textbf{(Successful Cases / Total Cases)}} & \makecell[c]{\textbf{Target Arrival Rate} \\ \textbf{(Successful Cases / Total Cases)}} \\
\midrule
Shortest Path (Dijkstra on vessel skeleton) & 12 / 20 (60.0\%) & 11 / 20 (55.0\%) \\
Heuristic-based (Topology features + heuristic rules) & 15 / 20 (75.0\%) & 14 / 20 (70.0\%) \\
\textbf{Proposed Method (Deep learning + GAT reasoning)} & \textbf{19 / 20 (95.0\%)} & \textbf{18 / 20 (90.0\%)} \\
\bottomrule
\end{tabular}
\end{table*}

\subsection{Performance Assessment of Vessel Segmentation}

For quantitative evaluation, the segmentation performance of six representative models was compared: U-Net, Attention U-Net, DeepLabv3+, TransUNet, Swin-UNet, and the proposed SCAR-UNet. The models were assessed using Dice coefficient, Precision, Recall, and Frames Per Second (FPS), as shown in Table~\ref{tab:seg}.

The results indicate that SCAR-UNet achieved the best overall performance, with a Dice coefficient of 93.1\%, Precision of 92.0\%, and Recall of 94.5\%. Compared to U-Net, both Attention U-Net and DeepLabv3+ provided moderate improvements in accuracy while maintaining relatively high inference speed. Transformer-based approaches, such as TransUNet and Swin-UNet, demonstrated higher segmentation accuracy than CNN-based baselines, with Dice scores of 91.6\% and 92.0\%, respectively, but suffered from lower inference speeds (49 FPS and 45 FPS). In contrast, SCAR-UNet not only surpassed all baseline methods in accuracy but also achieved a favorable trade-off between segmentation precision and computational efficiency. 

An ablation study was conducted by removing one component at a time, including CoordAttention, SimAM, SE, ASPP, and replacing the hybrid loss with Dice-only. As shown in Table~\ref{tab:ablation}, eliminating any attention mechanism led to a reduction in Dice, highlighting their complementary role in preserving thin vessels. Removing ASPP caused the largest Recall drop, underscoring the value of multi-scale context, while replacing the hybrid loss decreased both Dice and Recall, confirming the effectiveness of the Focal Tversky term in handling class imbalance. The complete SCAR-UNet achieved the highest accuracy while maintaining near real-time speed (58 FPS).

\subsection{Validation on Clinical DSA Images}
Path planning on DSA images plays a critical role in robotic interventions, particularly when apparent vessel crossings are present. A comparative study was conducted among the conventional shortest-path method, a heuristic-based approach, and the proposed framework, as illustrated in Figure~\ref{fig:Result}(a).

In all three trials, the shortest-path method consistently failed, as it only minimized path length without considering anatomical plausibility. In Trials 1 and 2, where the crossing angles were relatively large, the heuristic-based method was able to identify the correct branch. However, in Trial 3, the crossing angle was smaller and the heuristic rules were no longer sufficient, leading to an incorrect path. In contrast, the proposed method successfully resolved all three scenarios. By leveraging extensive annotated training data, integrating deep learning-based segmentation with GAT reasoning, and incorporating DSA patch features at crossing points, the proposed approach achieved robust disambiguation of apparent bifurcations and anatomically accurate path planning.

As summarized in Table~\ref{tab:comparison}
, the proposed method attained a disambiguation success rate of 95\% and a target arrival rate of 90\%, which substantially outperformed both the shortest-path method (60\% and 55\%) and the heuristic-based approach (75\% and 70\%). These results demonstrate that the integration of topological reasoning with local image features provides a more reliable and generalizable solution for robotic vascular navigation.

\subsection{Validation on Robotic System with Monocular Camera}

To further evaluate the framework in a robotic environment, experiments were conducted on a vascular phantom platform equipped with a robotic arm, gripper, pulsatile pump, and a monocular camera, that provided DSA-like vascular images in real time, as shown in Figure~\ref{fig:Result}(b).

As shown in Figure~\ref{fig:Result}(c), both the conventional shortest-path method and the heuristic-based approach were unable to generate correct trajectories on complex vascular crossings. The shortest-path method showed poor performance since it relied solely on minimizing path length without accounting for anatomical topology. The heuristic-based approach was misled at apparent crossing nodes where angle differences and vessel diameters were subtle, making its rule-based strategy unreliable. In contrast, the proposed SCAR-UNet-GAT framework successfully handled these challenging scenarios. By incorporating image-level features, combining deep learning-based segmentation with graph attention reasoning, and exploiting local patches at crossing points, the framework provided robust disambiguation of projection-induced ambiguities and produced anatomically plausible trajectories.

Building upon these results, Figure~\ref{fig:Result}(d) presents validation on the robotic system using the planned trajectories from Figure~\ref{fig:Result}(c). During the experiment, guidewire tip positions were detected in real time by a trained YOLOv5 model, and PID-based control was employed to regulate advancement and rotation of the gripper. By continuously correcting deviations between the detected tip and the reference trajectory, the guidewire was able to follow the planned path without executing unnecessary turns at false bifurcations. This closed-loop tracking reduced the likelihood of vessel wall contact and provided a safer and more reliable path-following strategy under robotic assistance.

\section{Discussion}
The SCAR-UNet-GAT framework achieved superior performance in coronary artery DSA segmentation and path planning compared with conventional approaches. SCAR-UNet attained a Dice coefficient of 93.1\%, providing a reliable foundation for centerline extraction. By integrating topological features with graph attention reasoning, projection-induced ambiguities were substantially reduced, with disambiguation and target-reach rates of 95.0\% and 90.0\%, respectively. Conventional shortest-path planning consistently failed at ambiguous crossings, whereas heuristic-based methods were only effective under specific geometric conditions. 

In addition to offline validation, the framework was further evaluated on a robotic platform, where it demonstrated the ability to generate anatomically consistent paths and guidewire navigation consistent with vascular topology. These results highlight the feasibility of integrating perception, planning, and control, thereby advancing toward closed-loop robotic-assisted interventions. 

Future-facing extensions can further strengthen clinical readiness and scope. Broader multi-center and multi-modality validation would better capture inter-patient and inter-scanner variability, and incorporating multi-view or 3D vascular context could mitigate residual ambiguities at complex junctions inherent to 2D projections. Additional priorities include uncertainty-aware graph reasoning with reliability estimation to temper the influence of upstream segmentation and skeletonization variability, and augmenting PID tracking with adaptive or learning-based control to enable resilient closed-loop navigation in tortuous or dynamically evolving vessels.

\section{Conclusion and Future Work}
This paper introduced a unified SCAR-UNet-GAT framework for vessel segmentation, centerline extraction, and topology-aware route inference under 2D DSA. By coupling multiscale attention–augmented segmentation with graph attention over topology-encoded geometric and appearance features, the approach reliably disambiguates projection-induced crossings and selects anatomically plausible paths. Validation on clinical DSA data and a benchtop robot platform demonstrated consistent gains over shortest-path and heuristic baselines in crossing disambiguation and target arrival, and confirmed the feasibility of integration within robot-assisted endovascular navigation workflows.

Future work will focus on closing the perception, planning, and control loop and broadening clinical applicability. Priorities include uncertainty-aware graph inference with reliability estimation to flag ambiguous crossings, incorporation of multi-view or 3D vascular context to mitigate projection artifacts, and learning-based motion planning and reinforcement learning for robust closed-loop guidewire control under imaging variability. Broader evaluation across multi-centre and multi-modality datasets, together with domain adaptation and calibration strategies, will be pursued to strengthen generalization and clinical reliability. These developments aim to further improve safety, robustness, and readiness for deployment in robot-assisted endovascular interventions.

\bibliographystyle{ieeetr}
\balance
\bibliography{reference}
\end{document}